\documentclass[letterpaper, 10 pt, conference]{ieeeconf}  

\IEEEoverridecommandlockouts                              

\usepackage[noadjust]{cite}

\usepackage{graphics} 
\usepackage{epsfig} 
\usepackage{xcolor}

\title{\LARGE \bf
BPActuators: Lightweight and Low-Cost Soft Actuators by Balloons and Plastics 
}

\author{Qiukai Qi$^{1}$, Shogo Yoshida$^{2}$, Genki Kakihana$^{2}$, Takuma Torii$^{2}$, Van Anh Ho$^{2}$ and Haoran Xie$^{2*}$
\thanks{This work was supported in part by JSPS KAKENHI under Grant JP20K19845 and 20J15087, and in part by Hayao Nakayama Foundation Research Grant A. Qiukai Qi is also supported by JSPS Doctoral Research Fellowship (DC2). \textit{(Corresponding author: Haoran Xie)}}
\thanks{$^{1}$Qiukai Qi was with the Graduate School of Advanced Science and Technology, Japan Advanced Institute of Science and Technology, and is now with the Department of Mechanical Engineering, Tokyo Institute of Technology, Tokyo, 152\,--\,8552, Japan
        {\tt\small qiukai.qi715@gmail.com}}%
\thanks{$^{2}$Shogo Yoshida, Genki Kakihana, Takuma Torii, Van Anh Ho and Haoran Xie are with the Graduate School of Advanced Science and Technology, Japan Advanced Institute of Science and Technology,
        Ishikawa, 923\,--\,1292, Japan
        {\tt\small xie@jaist.ac.jp}}%
}

\begin{document}

\maketitle
\thispagestyle{empty}
\pagestyle{empty}

\begin{abstract}

To increase the awareness and impact, soft robotics needs to go beyond the lab environment and should be readily accessible to those even with no robotic expertise. However, most prevailing manufacturing technologies require either professional equipment or materials which usually are not available to common people, thereby constraining the accessibility. In this communication, we propose a lightweight and low-cost soft bending actuator, called ``BPActuator", that can be easily manufactured with Balloons and Plastics. For characterization, we fabricated a range of actuators with different morphology and tested in terms of the capability of deformation and load-bearing. We demonstrated that they can bend up to 35 degrees and exert force at the tip around 0.070\,$\pm$\,0.015\,N, which is over 5 times higher than their average gravity. We further implemented a gripper of three fingers using BPActuators, and found that the gripper can realize human-like grasp of a range of daily objects. To be specific, the gripper can lift objects at least 8 times heavier than its own weight. Moreover, BPActuators are cost effective and each costs about 0.22\,USD. Given all the advantages, we believe that the BPActuators will significantly improve the accessibility of soft robotics.

\end{abstract}

\section{INTRODUCTION}

Soft robotics has drawn a growing interest from a wide group spanning from medical and industry to entertainment and education in the last decade \cite{sororeview_daniela, sororeview_cecilia, sororeview_jonathan, soroeducation}. They are typically compliant, adaptable, and thereby secure to interact with surrounding environment, making soft robotics a promising solution for tasks characterized by human machine interaction \cite{sororeview_jonathan}. Despite the numerous successful demonstrations such as soft grippers \cite{softgripper} and sensors \cite{softsensor1, softsensor2}, soft robotics has remained largely in the laboratory environment, limiting the accessibility of soft robotics to those who have no robotic expertise. To further increase the ubiquity of soft robots, they are required to be more readily available to a wider group of people. In this perspective, soft robotics is preferred to be more lightweight, low-cost and easy-to-fabricate \cite{soropathahead}.

The development of soft robotics has been greatly improved by the advancement of manufacturing technology and material science \cite{soropathahead}. Many technologies for manufacture have been investigated for fabrication of soft robots, including, for example, 3D printing, shape deposition manufacturing and soft lithography \cite{sororeview_daniela, fludic_review}. While these technologies have enabled the rapid and effective prototyping of soft robotics with complex structures, they also pose challenges to further boost the accessibility since  they all require professional equipment or materials that are not available to common people. A broad range of materials have been adopted to fabricate soft actuators, which includes, for example, hydrogels, silicon rubbers, textiles and plastic materials (polyethylene) \cite{fludic_review, soro3dprintingmaterial}. Among these options, the plastic materials have been popular because of their availability and effective cost. They have been explored in various interesting works, such as sPAM \cite{spam}, pouch motors \cite{pouchmotors}, bubble actuators \cite{softbubbleactuator}, BlowFab \cite{blowfab} and balloon bending actuators \cite{balloon1}. One concern, however, is that the requirement for professional equipment, such as printing device \cite{pouchmotors} and heat bonding device \cite{balloon1}, and the special precautions to make airtight chambers may complicate the fabrication process.

\begin{figure}[t]
    \centering
        \includegraphics[keepaspectratio=true,width=85mm]{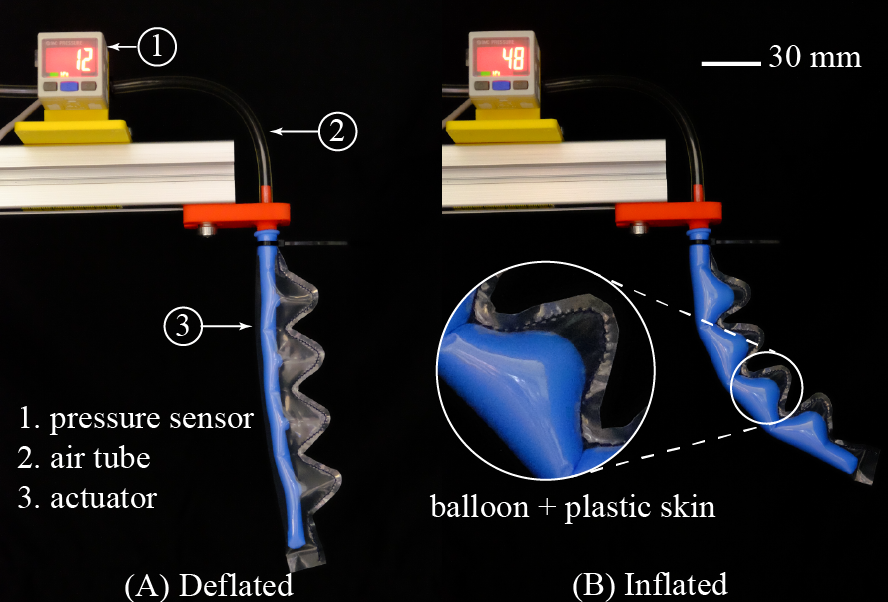}
            \caption{The soft actuator under de- and inflated states. (A) is the deflated state, and (B) is the inflated state. The actuator consists of a balloon, as air-tight chamber, covered by a plastic skin, with stitched sinusoidal rim, constraining the buckling of the internal balloon.}
            \label{first}
\end{figure}

In this communication, we propose a soft bending actuator as illustrated in Fig. \ref{first}, named ``BPActuator". The actuator can be fabricated simply by inserting a balloon into a plastic skin that was cut and stitched or stapled to a wavy shape as shown in Fig. \ref{first}. When inflated, the balloon buckles and extends into the wave, triggering the bending deformation. Here, all materials are readily available to those who are interested, and the fabrication requires no prior robotic knowledge. Here, we use balloons to form the air chamber, eliminating the need for the complex fabrication of airtight plastic skins. For characterization, we fabricated a series of actuators with different wavelengths to verify the capacity of deformation and load bearing, and finally implemented a robotic gripper to demonstrate the feasibility of such design for soft robotics. This low-cost (0.22\,USD/unit estimated) and easy-to-fabricate design, we believe, will contribute to bringing soft robotics beyond the lab environment, greatly improving their accessibility to a broader group.

\section{FABRICATION, DESIGN AND WORKING PRINCIPLE}
In this section, the fabrication process of BPActuator is first introduced, then followed by the design and working principle for better understanding.

\begin{figure}[t]
    \centering
        \includegraphics[keepaspectratio=true,width=85mm]{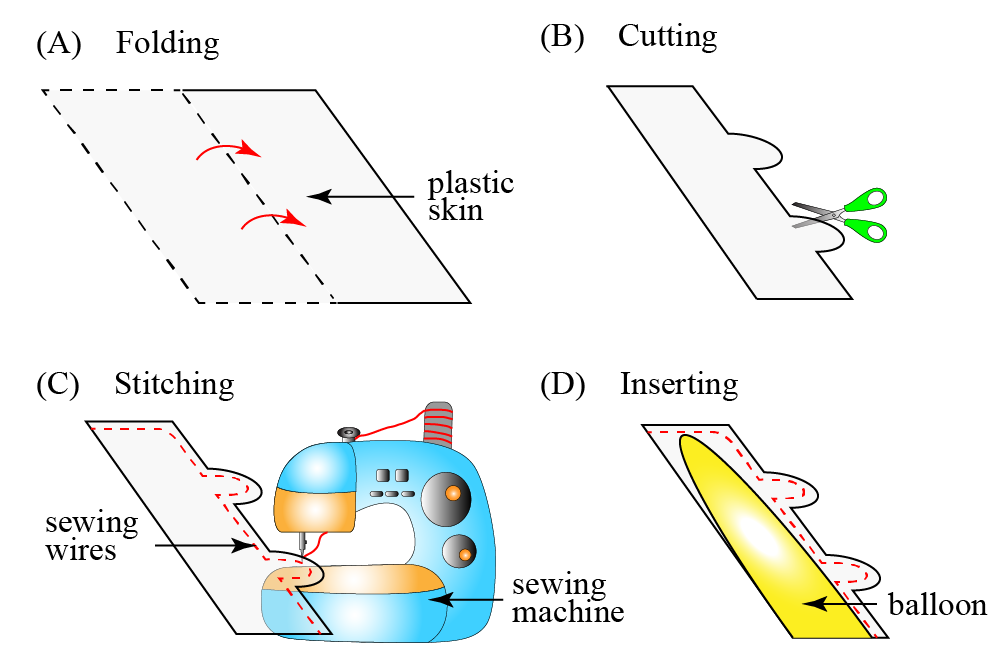}
            \caption{The fabrication process of the actuator. (A) Fold the plastic sheet symmetrically along the central axis. (B) Cut the plastic skin to the desired shape, giving the actuator skin. (C) Stitch the rim either manually or with a sewing machine. (D) Insert the balloon into the plastic skin and spread it evenly.}
            \label{fabrication}
\end{figure}

\subsection{Fabrication}
The BPActuator was fabricated by inserting a balloon into an inextensible but flexible plastic skin, as illustrated in Fig. \ref{fabrication}. In detail, the plastic skin (material: polyethylene; thickness: 0.08\,mm) was first folded along the central line, and cut to shape as designed. Then it was stitched along the rim either manually or by a sewing machine or simply by a stapler. Finally, the balloon (material: natural latex rubber; dimension: around $\Phi$50$\times$1500\,mm when inflated) was inserted to complete the process. Here, both materials can be easily obtained at very low cost (balloon: 0.011\,USD/unit; plastic: 0.21\,USD/unit).

\subsection{Design}
The detailed structural design is presented in Fig. \ref{design}. The contracting side was designed to be with a sinusoidal boundary. As shown in Fig. \ref{design} (A), all dimensional parameters are fixed except the wavelength (b) that ranges from 25\,mm to 40\,mm with an increment of 5\,mm, yielding 4 different morphology totally (Note that the wavelength could be either 3 or 4 because of the fixed total length). It was so because we intended to investigate how the wavelength is to affect the performance.

\subsection{Working Principle}
\begin{figure}[t]
    \centering
        \includegraphics[keepaspectratio=true,width=85mm]{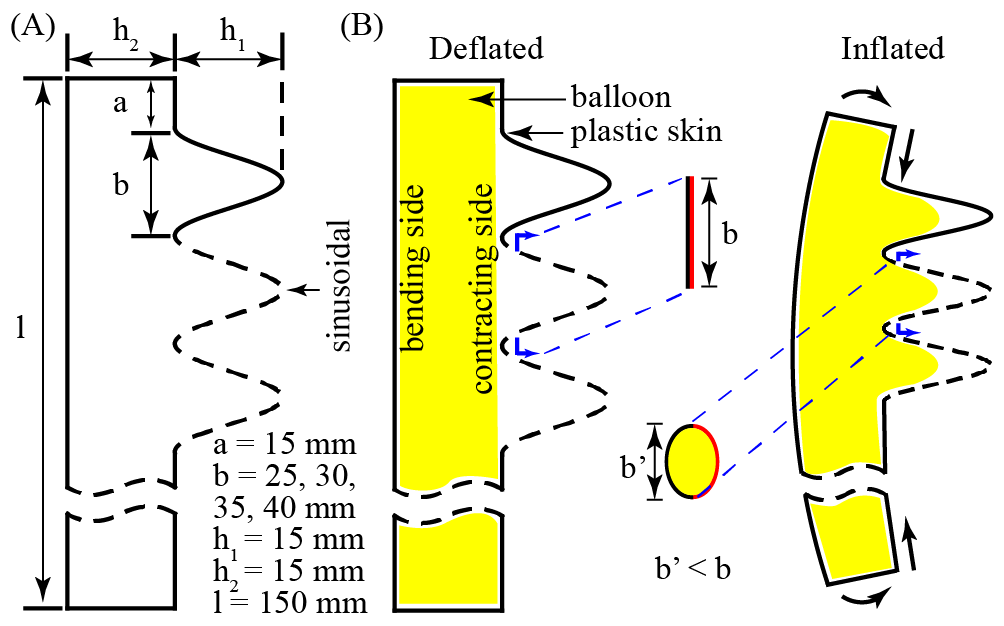}
            \caption{The structural design and working principle of the actuators. In (A), all parameters are fixed and indicated as shown except the wavelength that varies from 25\,mm to 40\,mm with a 5\,mm increment, yielding four actuator designs. (B) illustrates the bending principle due to the asymmetric lateral contraction in length caused by the buckling balloon.}
            \label{design}
\end{figure}

Fig. \ref{design} (B) presents a simple understanding of the working principle of these actuators. The actuator is initially flat, which can be conceived from the fabrication. Take a closer look at one wave bump as in Fig. \ref{design} (B), the wavelength is denoted by b. The black and red lines indicate the two layers of plastic skin, respectively. When gradually inflated, the balloon buckles and extends into the wavy chamber that can be studied by the elliptical cross section as in Fig. \ref{design} (B). It is seen that length of the major axis contracts to the current length (b') that is shorter than its initial state (b). The accumulation of the shortening waves constitute the contracting side of the actuators, driving the actuators to bend, marked as in Fig. \ref{design} (B).

\section{EXPERIMENTS AND RESULTS}

The capabilities of deformation and load-carrying are two important factors for actuators. In this section, the BPActuators with different wavelengths are characterized in terms of these two aspects.

\subsection{Deformation Capability}
\subsubsection{Method}
The BPActuators were inflated to different pressure levels (10, 20, 30, 40, 50\,kPa) modulated by a pressure regulator (IVT0030-2CL, SMC, Japan) controlled by Python via a data acquisition device with analog output functionality (NI USB-6343, National Instrument, USA). A camera was used to capture the deformation. The bending angle marked in Fig. \ref{deformation} was then extracted by processing the images with OpenCV (https://opencv.org/) in Python.

\subsubsection{Results}
\begin{figure}[t]
    \centering
        \includegraphics[keepaspectratio=true,width=85mm]{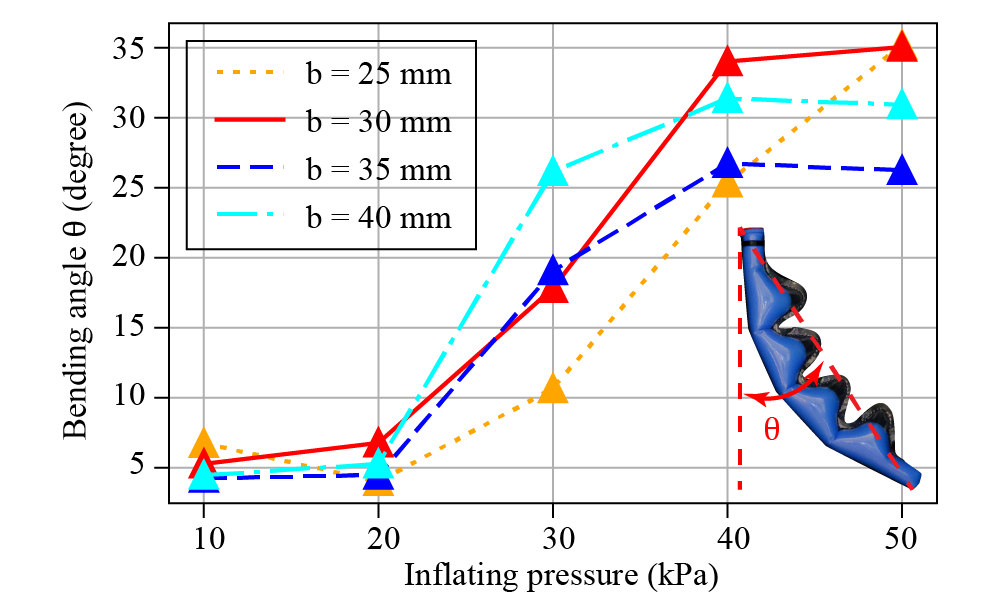}
            \caption{The characterization of deforming capability of the actuators. The inset indicates the bending angle definition. Here, only the design with four different wavelengths are presented. Note that there can be 3 or 4 waves on one actuator depending on the different wavelengths.}
            \label{deformation}
\end{figure}

The bending angle of actuators with four different wave length designs are presented in Fig. \ref{deformation}.  The inflating balloons in all four designs, as can be seen, first fill the main body of the actuator, then buckle and extend into the wavy chambers starting from 20\,kPa. When fully extended, the actuators stop bending further significantly even with higher pressure. The stopping pressure is different depending on the wavelength. As indicated in Fig. \ref{deformation}, the shorter the wavelength  is, the higher pressure is needed to fully extend.

\subsection{Load-bearing Capability}
\subsubsection{Method}
The force was measured by a loadcell (Nano 17, ATI
Industrial Automation, USA) mounted on a X-Z linear stage (KXL 06150, KZL 06075, Suruga Seiki, Japan), as shown in Fig. \ref{force} (A). The loadcell was controlled to press towards the actuator tip by 2\,mm and measurement was recorded via the data acquisition device in the meantime. Each measuring was conducted 3 times. Note that the load bearing property is dependent on the contact locations, therefore we select the actuator tip for generality.

\subsubsection{Results}
\begin{figure}[t]
    \centering
        \includegraphics[keepaspectratio=true,width=85mm]{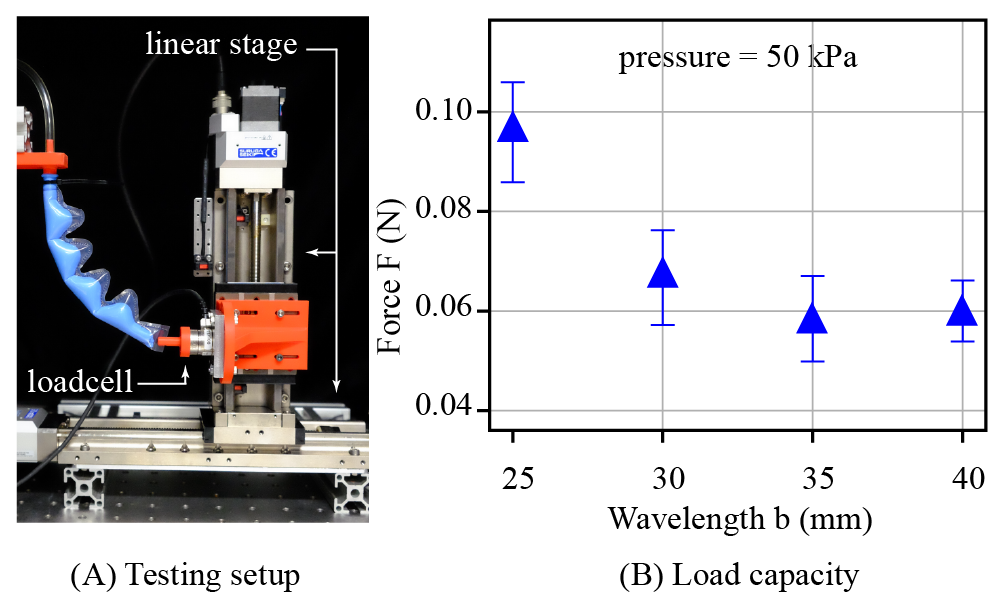}
            \caption{The characterization of load bearing capability of the actuators. (A) indicates the experimental setup which includes a loadcell, for force measurement, and X-Z linear stage for loadcell movement. (B) presents the force detected when pushing by 2\,mm towards the actuator tips with 50\,kPa inflating pressure.}
            \label{force}
\end{figure}

The mean and standard deviation of the force reading of each design under 50\,kPa are presented in Fig. \ref{force} (B). In average, the load bearing capacity of these actuators is 0.070\,$\pm$\,0.015\,N that is over 5 times higher than their average gravity (1.392\,$\pm$\,0.043\,g\,$\times$\,0.0098\,N/g). Further, the shorter wavelength design tends to behave better. This is considered due to the more evenly distributed buckling points.

\begin{figure}[t]
    \centering
        \includegraphics[keepaspectratio=true,width=85mm]{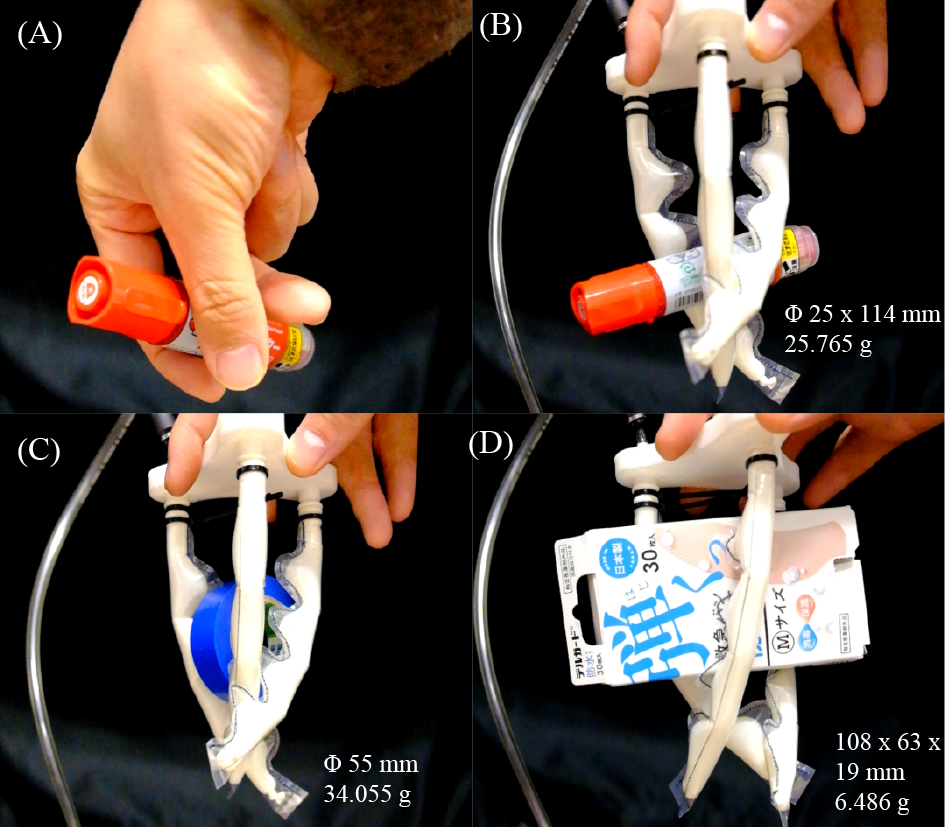}
            \caption{The demonstration of a typical human grasping posture with three fingers in (A) and of the mimicking robotic grasping of various objects including a marker pen (B), a tape (C) and a pack of bondage (D). The weight of each object is labeled accordingly.}
            \label{application}
\end{figure}

\section{Application}
A robotic gripper with three fingers was developed with the BPActuators, as illustrated in Fig. \ref{application}, to demonstrate the applicability of these designs. Specifically, three actuators with wavelength of 30\,mm were mounted on a 3D printed gripper palm. When actuated, the gripper can realize the human-like grasping posture, as shown in Fig. \ref{application} (A), and is capable of grasping a wide range of objects including a marker pen (25.765\,g) (B), a rubber tape (35.055\,g) (C) and a pack of bondage (6.486\,g) (D). It was demonstrated that the gripper can grasp objects at least 8 times more than the own weight. The real time grasping is also shown in the supplementary movie \textbf{S1}.

\section{CONCLUSIONS}

In this communication, we reported a lightweight, low-cost and easy-to-fabricate actuator design (``BPActuator") and verified the effectiveness in terms of deformation and load-bearing capability. We further revealed the potential of this design as a soft robotic gripper capable of grasping and holding various objects that can be at least 8 times heavier than the own weight. For future plans, we will model the morphing behavior in order to predict the movement for better control. And we will implement multiple applications, for example toy driving, to reveal the potential of BPActuators.

\addtolength{\textheight}{-12cm}   



\section*{ACKNOWLEDGMENT}

The authors would like to thank Prof. Koichi Suzumori for valuable comments, Haruka Suzuki for his assistance in gripper demonstration, and Aoshi Chiba and Ping Zhang for their involvement in the initial actuator design.

\end{document}